\documentclass[letterpaper, 10 pt, conference]{ieeeconf}  % Comment this line out if you need a4paper

\IEEEoverridecommandlockouts                              % This command is only needed if 
\usepackage{graphicx}
\usepackage{hyperref}
\newcommand{\linkToPdf}[1]{\href{#1}{PDF}}
\usepackage{balance}
\usepackage{tabularx}
\usepackage{multirow}
\usepackage{multicol}
\usepackage{booktabs}
\usepackage{mathtools}
\usepackage{amssymb}
\usepackage{siunitx}
\usepackage{amsmath}
\usepackage{MnSymbol}
\usepackage{kotex}
\usepackage{tablefootnote}
\usepackage{bbding}
\usepackage{threeparttable} 
\usepackage{cite}
\hypersetup{
    colorlinks=False, 
    urlcolor=black,
}

\usepackage{kotex}
\usepackage{color}
\usepackage{xcolor}

\title{\LARGE \bf
VIGS SLAM: IMU-based Large-Scale 3D Gaussian Splatting SLAM
}

\author{Gyuhyeon Pak$^{1}$ and Euntai Kim$^{1, *}$% <-this % stops a space
\thanks{$^*$ is that the corresponding author.}
\thanks{$^{1}$Gyuhyeon Pak and Euntai Kim are with the Department of Electrical and Electronic
Engineering, Yonsei University, Seoul 03722, South Korea
        {\tt\small \{gh.pak, etkim\}@yonsei.ac.kr}}%
}

\begin{document}

\maketitle
\thispagestyle{empty}
\pagestyle{empty}

\begin{abstract}
Recently, map representations based on radiance fields such as 3D Gaussian Splatting and NeRF, which excellent for realistic depiction, have attracted considerable attention, leading to attempts to combine them with SLAM. 
While these approaches can build highly realistic maps, large-scale SLAM still remains a challenge because they require a large number of Gaussian images for mapping and adjacent images as keyframes for tracking.
We propose a novel 3D Gaussian Splatting SLAM method, VIGS SLAM, that utilizes sensor fusion of RGB-D and IMU sensors for large-scale indoor environments.
To reduce the computational load of 3DGS-based tracking, we adopt an ICP-based tracking framework that combines IMU preintegration to provide a good initial guess for accurate pose estimation.
Our proposed method is the first to propose that Gaussian Splatting-based SLAM can be effectively performed in large-scale environments by integrating IMU sensor measurements. 
This proposal not only enhances the performance of Gaussian Splatting SLAM beyond room-scale scenarios but also achieves SLAM performance comparable to state-of-the-art methods in large-scale indoor environments.

\end{abstract}

%%%%%%%%%%%%%%%%%%%%%%%%%%%%%%%%%%%%%%%%%%%%%%%%%%%%%%%%%%%%%%%%%%%%%%%%%%%%%%%%
\section{INTRODUCTION}

Simultaneous Localization and Mapping (SLAM) is a critical technology in the fields of robotics and autonomous systems, aiming to simultaneously solve the problems of estimating a robot's location and constructing a map of the surrounding environment. This task is essentially challenging due to the interdependence between localization and mapping. Accurate map construction requires precise localization, and conversely, precise localization depends on an accurate map.

Recently, studies that utilize the concept of radiance fields, such as NeRF\cite{mildenhall2021nerf} or 3D Gaussian Splatting (3DGS)\cite{kerbl20233dgs}, to represent 3D spaces have received significant attention due to their advantages in realistic depiction and fast rendering speeds.
These advantages have led to new approaches that combine radiance field-based methods with SLAM\cite{rosinol2023nerfslam, johari2023eslam, keetha2024splatam, matsuki2024monogs, yan2024gsslam, huang2024photoslam, sun2024mm3dgs, ha2024gsicpslam}, to improve the accuracy and efficiency of both localization and mapping.
In particular, 3DGS processes data by considering the uncertainty at each point, allowing for the construction of highly accurate and realistic maps. 
As a result, 3DGS-based SLAM\cite{keetha2024splatam, matsuki2024monogs, yan2024gsslam, huang2024photoslam, ha2024gsicpslam, sun2024mm3dgs} (breifly, 3DGS SLAM) has emerged as a promising technology for efficiently handling high-resolution visual data, thereby improving robot localization and environmental understanding.

\begin{figure}[t]
\centering
\includegraphics[width=0.98\columnwidth]{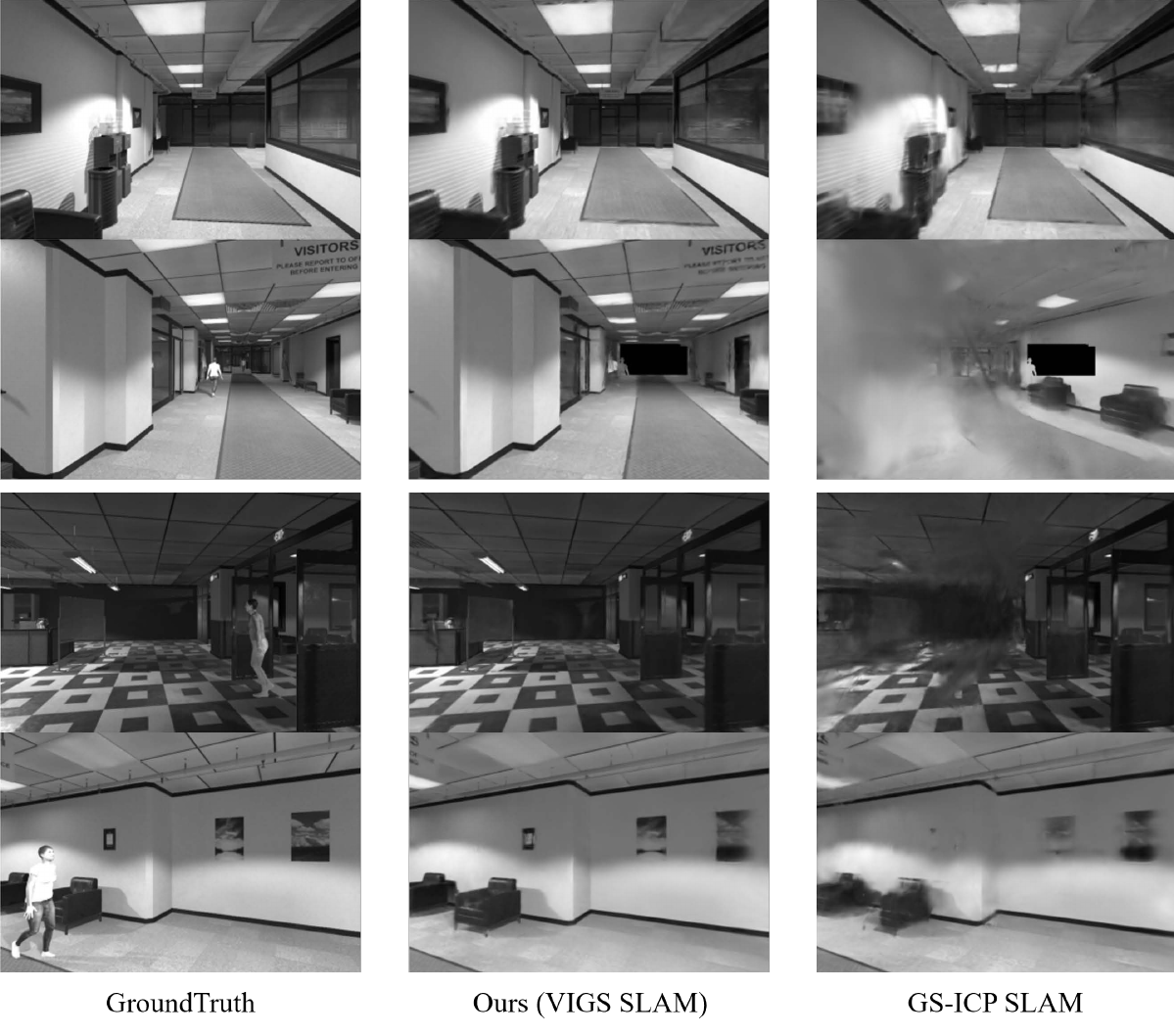}
\caption{Comparison of rendering image quality according to initial estimates on Humans12 (upper 2 rows) and Humans24 (lower 2 rows) of uHumansV1 dataset}
\label{fig_rendering_image_uhumansv1}
\end{figure}

Basically, 3DGS SLAM \cite{keetha2024splatam, matsuki2024monogs, yan2024gsslam, huang2024photoslam, ha2024gsicpslam, sun2024mm3dgs} consists of a front-end and a back-end: The front-end is responsible for tracking and localization while the back-end is responsible for optimization and map construction. This paper focuses on the front-end of 3DGS SLAM. Reflecting broader trends in classical SLAM research, there are generally two main approaches in the front-end of 3DGS SLAM: the direct method and the feature-based method.

The first approach, the direct method\cite{keetha2024splatam, yan2024gsslam, matsuki2024monogs}, conducts tracking by comparing the raw sensor image with the reconstructed images using dense photometric loss. Tracking in the direct method is easily combined with back-end optimization (i.e., optimization for 3DGS reconstruction), leading to high accuracy. 
However, this method requires that keyframes are relatively closely spaced due to the sensitivity of the dense photometric loss, thereby storing a large number of keyframes. As a result, the direct method has difficulty in being applied to large-scale environments due to memory limitations.

The second approach is the feature-based method\cite{huang2024photoslam}. Since it stores only a limited number of features, it requires significantly less memory to store the map, making it suitable for large-scale environments. However, in feature-based 3DGS SLAM, the front-end tracking and back-end optimization are mostly decoupled, which results in a notable degradation in accuracy. In summary, neither method performs well in terms of both accuracy and scalability to large environments.

To address this challenge, we propose a novel 3DGS SLAM approach called Visual-Inertial Gaussian Splatting (VIGS) SLAM. Our SLAM leverages sensor fusion between an RGB-D sensor and an Inertial Measurement Unit (IMU) for being applied to large-scale indoor environments. The RGB-D sensor provides both visual and depth information, enabling more accurate 3D environment perception, while the IMU captures data related to the robot's motion by measuring acceleration and angular velocity. By effectively fusing data from these two sensors, the performance of the SLAM system is significantly enhanced.

Combining IMU and 3DGS SLAM is not completely new, but a paper which applies IMU to direct method was very recently reported \cite{sun2024mm3dgs}. However, we believe that combining dense direct tracking with IMU is not the optimal solution.
This approach can only slightly increase the distance between keyframes (i.e., it can slightly reduce the number of keyframes) due to the dense nature of photometric loss, making it difficult to apply to large-scale environments. 

Instead, we propose applying IMU data not to raw image matching, but to the matching of the point cloud generated from the depth image. This new combination overcomes the limitations of dense photometric loss, significantly increasing the distance between keyframes (thereby significantly reducing the number of keyframes and memory usage) while maintaining the connection between the front-end and back-end, thus preserving high accuracy. Thus, we believe that our method can achieve both high accuracy and large-scale coverage. Our approach uses \cite{ha2024gsicpslam} as a base line and has the following contributions.

\begin{itemize}
\item We propose a large-scale visual-inertial SLAM framework, VIGS SLAM, leveraging sensor fusion between RGB-D and IMU sensor, inspired by GS-ICP SLAM \cite{ha2024gsicpslam}.
\item By utilizing the IMU preintegration values between consecutive frames, we improve the accuracy of the initial guesses. These good initial guesses play a crucial role in improving the accuracy of both pose estimation and mapping vy provinding more precise inputs to the Iterative Closest Point (ICP) algorithm.
\item We have developed a SLAM system that outperforms existing room-scale 3DGS SLAM, enabling efficient operation in large-scale indoor environments. Our approach represents a significant advance in large-scale indoor SLAM, demonstrating both scalability and robustness comparable to state-of-the-art SLAM systems.
\end{itemize}

The remainder of this paper is organized as follows: 
In Section \ref{related works}, we review related works on SLAM leveraging multiple sensor fusion and 3D GS SLAM. In Section \ref{method}, we provide a detailed description of the proposed method, including Generalized ICP tracking, IMU preintegration, 3D Gaussian Splatting mapping. Section \ref{experiments} presents experimental results demonstrating the qualitative and quantitative performance of the proposed method in large-scale indoor environments. Finally, in Section \ref{conclusions}, we discuss the conclusions of the proposed method and potential directions for future research.

\section{RELATED WORKS}
\label{related works}

\subsection{Sensor Fusion for SLAM}
Although visual perception through camera sensors is effective, it faces limitations such as motion blur, changes in exposure, and sensitivity to lighting conditions. One way to overcome these limitations is through the fusion of complementary sensors, which compensates for the weaknesses of each sensor. In this regard, previous approaches have enhanced SLAM performance by integrating depth sensors\cite{whelan2013robust, mur2017orb}, LiDAR\cite{hong2024livgaussmap, shan2021lvisam, lang2024gaussianlic}, and IMU sensors\cite{qin2018vins,geneva2020openvins,forster2015imu, Rosinol21ijrr-Kimera_uhumans2} with visual SLAM, thereby addressing issues like scale ambiguity and improving the robustness of SLAM systems. 
In this paper, we focus on the fusion with IMU sensors, which are particularly effective for their ability to track rapid movements within short time frames and their high data acquisition rate, making them ideal for integrating inertial measurements.

\begin{figure*}[ht]
\centering
\includegraphics[width=\textwidth]{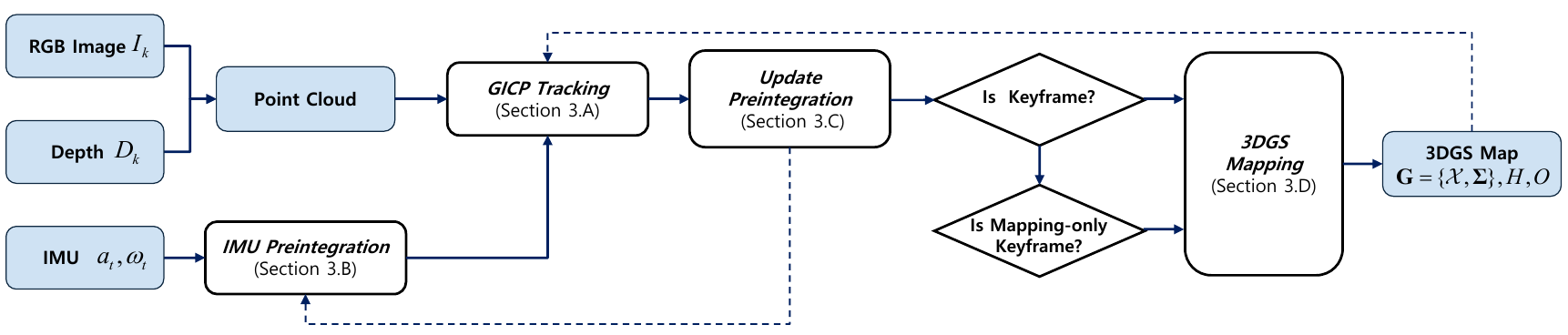}
\caption{ \textbf{Overview of VIGS SLAM}. The input of VIGS SLAM is RGB-D image and IMU meausrements. The system generates a point cloud from RGB and depth inputs, followed by GICP tracking. The IMU preintegration values are used as a good initial guess to enhance tracking performance, and these values are updated after tracking each frame. Keyframes are identified for 3D Gaussian Splatting-based mapping, which efficiently updates the 3DGS map with detailed environmental representations. }
\label{fig_pipeline}
\end{figure*}

\subsection{3DGS SLAM}

3D Gaussian Splatting (3DGS) based methods \cite{keetha2024splatam, matsuki2024monogs, yan2024gsslam} have detailed the significant advantages of 3DGS over traditional map representations in SLAM tasks for online photo-realistic mapping. 
Additionally, they highlight that the vanilla 3DGS must be appropriately adapted for efficient application in SLAM. 
SplaTAM\cite{keetha2024splatam} and GS-SLAM\cite{yan2024gsslam} describe the benefits of 3DGS map representations over conventional SLAM map representations. 
MonoGS\cite{matsuki2024monogs}, using a single RGB sensor, addresses the ambiguity of incremental reconstruction by introducing geometric regularization, making it work well with monocular images. 

In contrast, methods such as GS-ICP SLAM\cite{ha2024gsicpslam} and Photo-SLAM\cite{huang2024photoslam} estimate the necessary poses for mapping using classical visual odometry to estimate camera movement, thereby enabling sufficiently fast SLAM operations. 
Photo-SLAM\cite{huang2024photoslam} leverages the classical visual odometry method, ORB-SLAM3\cite{campos2021orb}, for accurate pose estimation and reconstructs a hybrid Gaussian map that incorporates ORB features.
GS-ICP SLAM\cite{ha2024gsicpslam} performs pose tracking between successive point clouds using ICP-based tracking \cite{koide2021GICP} results and incorporates the covariance of each point, obtained during tracking, into 3DGS mapping to achieve real-time SLAM. 

\section{METHOD}
\label{method}

The VIGS SLAM framework consists of three main stages: 1) Generalized ICP tracking, 2) IMU preintegration, 3) 3D Gaussian Splatting mapping. An overview of the framework is described in Fig. \ref{fig_pipeline}.

\subsection{Generalized ICP Tracking}
\label{GICP tracking}
Let us suppose that the RGB image $I_{k}$ and depth image $D_{k}$ are presented at the $k$th frame. The corresponding point cloud $\boldsymbol{\mathcal{X}}_{k} = \{x_{m} \}_{{m=1,...,M_k}}$ is obtained from $D_{k}$, and the associated covariance set $\boldsymbol{\Sigma}_k = \{\Sigma_{m}\}_{{m=1,...,M_k}}$ is given by computing the covariance matrix of $k$-nearest neighbors of each point $x_{m}$, where  $x_{m} \in {\mathbb {R}}^3 $ and $\Sigma_m \in {\mathbb {R}}^{3 \times 3}$ are a 3D point  and the corresponding covariance matrix, respectively, and $m$ is an index for $\boldsymbol{\mathcal{X}}_{k}$ and $\boldsymbol{\Sigma}_k$.
Following the process, we can obtain the set of Gaussians $\boldsymbol{G}_{k}= \{\boldsymbol{\mathcal{X}}_{k}, \boldsymbol{\Sigma}_{k}\}$ corresponding to the point cloud at each frame.

The relative transformation $\boldsymbol{T}_{k}$ between the source Gaussians $\boldsymbol{G}_{k}^{src}= \{\boldsymbol{\mathcal{X}}_{k}^{src}, \boldsymbol{\Sigma}_{k}^{src}\}$ generated from current RGB-D image $\{I_{k}$, $D_{k}\}$ and corresponding target Gaussian set $\boldsymbol{G}_{k}^{tgt} = \{\boldsymbol{\mathcal{X}}_{k}^{tgt}, \boldsymbol{\Sigma}_{k}^{tgt}\}$ that constitutes the map $M$ can be estimated using Generalized ICP (GICP)\cite{koide2021GICP}.
By modeling each point $x_{m}$ as a Gaussian distribution $N(\hat{x}_{m}, \Sigma_{m})$, the distance $d_{m} = x^{tgt}_{m} - \boldsymbol{T}_{k}x^{src}_{m}$ between the corresponding distributions pair in the GICP framework is given by
\begin{eqnarray}
&d_{m} \sim N(\hat{d}_{m}, \Sigma^{tgt}_{m} + \boldsymbol{T}_{k}\Sigma^{src}_{m}(\boldsymbol{T}_{k})^{T})\\
&=N( \hat{x}^{tgt}_{m}- \boldsymbol{T}_{k}\hat{x}^{src}_{m}, \Sigma^{tgt}_{m} + \boldsymbol{T}_{k}\Sigma^{src}_{m}(\boldsymbol{T}_{k})^{T}).
\end{eqnarray}

We derive the optimal transformation $\boldsymbol{T}_{k}^{*}$ by applying the maximum likelihood estimation method to the previously expressed distance function. The optimal transformation $\boldsymbol{T}_{k}^{*}$ is obtained by maximizing the likelihood function:
\begin{align}
\boldsymbol{T}_{k}^{*} &= \displaystyle \underset{\boldsymbol{T}_{k}}{\mathrm{argmax}}\prod_{m=1}^{N} p(d_{m}) = \underset{\boldsymbol{T}_{k}}{\mathrm{argmax}}\sum_{m=1}^N \mathrm{log} p(d_{m}) \\
&= \displaystyle \underset{\boldsymbol{T}_{k}}{\mathrm{argmax}}\sum_{m=1}^N d_{m}^{T}(\Sigma^{tgt}_{m} + \boldsymbol{T}_{k}\Sigma^{src}_{m}(\boldsymbol{T}_{k})^{T})^{-1}d_{m}. 
\end{align}

\begin{figure}[t]
% \centering
\includegraphics[width=1.0\columnwidth]{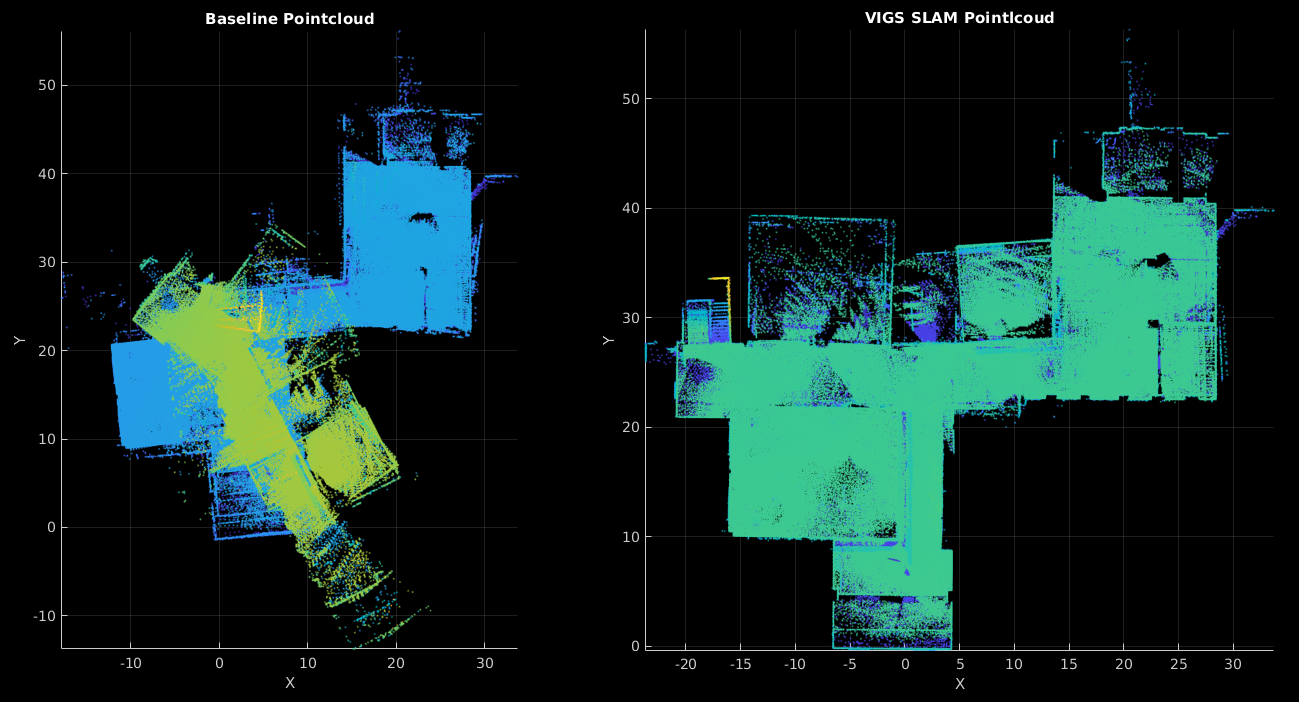}
\caption{Comparison of map reconstruction on uHumansV1. GS-ICP SLAM (left), VIGS SLAM (right)}
\label{fig_mapping_uhumansv1}
\end{figure}

GICP performs camera tracking by alternating between (1) finding the correspondences between the closest points, and (2) minimizing the Euclidean distance between these point pairs. In each iteration, the correspondences are recalculated, and the tracnsformation is updated through the optimization process. If the initial correspondences are accurate, even when the camera moves rapidly, GICP enables fast and reliable tracking, reducing mapping errors. The key idea of this paper is to use IMU preintegration as the intial estimate for the correspondences, allowing for efficient and reliable tracking. An example result is shown in Fig. \ref{fig_mapping_uhumansv1}. As illustrated, the mapping results of our baseline method are significanty skewed, whereas our VIGS SLAM demonstrates excellent mapping accuracy.

\subsection{IMU Preintegration}
The IMU sensor outputs linear acceleration $\hat{a}={[}\hat{a}_{x},\hat{a}_{y},\hat{a}_{z}{]}^{T}\in \mathbb{R}^{3}$ and angular velocity $\hat{\omega}=[\hat{\omega}_{x},\hat{\omega}_{y},\hat{\omega}_{z}]^{T} \in \mathbb{R}^{3}$ and helps measure the sensor's motion along these 6 degrees of freedom (DoF). Since the IMU operates at a much higer frequency than the camera, we pre-integrate the IMU sensor's measurements between RGB-D frames and use the pre-integration results as the initial guess for the sensor motion in GICP tracking part. In this paper we use k as the time index for RGB-D frames, whereas use t as the time index for the IMU sensor.

Since the measurements include inherent noise and bias of the sensor, the corrected linear acceleration $a_{t}$ and angular velocity $\omega_{t}$ are given by the following equations:
\begin{align}
&\hat{a}_{t} = a_{t} + \prescript{}{w}{R}\prescript{w}{}{g} + b_{a} + n_{a} \\
&\hat{\omega}_{t} = \omega_{t} + b_{\omega} + n_{\omega},
\end{align}
where $\prescript{}{w}{R}$ represents transpose of the sensor's rotation, $\prescript{w}{}{g}$ represents gravity vector, and $b_{a}, b_{g}, n_{a},$ and $n_{g}$ denote the biases and noises of the accelerometer and gyroscope, respectively.

Using the kinematic model\cite{forster2015imu}, the current frame's relative position can be predicted based on the IMU measurements.
\begin{align}
&p^{k}_{t+1} = p^{k}_{t} + v^{k}_{t}\Delta t + {\frac{1}{2}}(\hat{a}_{t} - \prescript{}{w}{R}\prescript{w}{}{g} - b_{a} - n_{a})\Delta t^{2} \label{eq7}\\
&v^{k}_{t+1}  = v^{k}_{t} + (\hat{a}_{t} - \prescript{}{w}{R}\prescript{w}{}{g} - b_{a} - n_{a})\Delta t \label{eq8}\\
&R^{k}_{t+1} = R^{k}_{t}\text{Exp}\{(\hat{\omega}_{t}-b_{\omega} -n_{\omega})\Delta t\},
\label{eq9}
\end{align}
where IMU sensor's position, velocity, and rotation are denoted as $p^{k}_{t}$, $v^{k}_{t}$, $R^{k}_{t}$, and $\Delta t$ is time difference between time $t$ and $t+1$.

To avoid repetitive computations for the motion between consecutive frames, the IMU sensor's preintegration\cite{qin2018vins} values can be expressed as follows:
\begin{align}
&\alpha^{k}_{k+1} = \iint_{t\in[t_{k}, t_{k+1}]} R^{k}_{t}(\hat{a}_{t} - b_{a})dt^{2} \\
&\beta^{k}_{k+1} =  \int_{t\in[t_{k}, t_{k+1}]}R^{k}_{t}(\hat{a}_{t} - b_{a})dt \\
&\gamma^{k}_{k+1} = \int_{t\in[t_{k}, t_{k+1}]}{\frac{1}{2}}\Omega(\hat{\omega}_{t}-b_{\omega})\gamma^{k}_{t} dt,
\end{align}
where
$\Omega(\omega) = \left[\begin{array}{cc}   {-\lfloor\omega\rfloor}_{\times} & \omega \\ -\omega^{T} & 0 \\ \end{array}\right],$ 
${\lfloor\omega\rfloor}_{\times} =  \begin{bmatrix}
 0 & -\omega_{z} & \omega_{y} \\ 
 \omega_{z} & 0 & -\omega_{x} \\ 
 -\omega_{y} & \omega_{x} & 0 \end{bmatrix}$ represents the skew-symmetric matrix associated with the angular velocity $\omega$.

Using (\ref{eq7}), (\ref{eq8}), and (\ref{eq9}), the relative transformation between consecutive frames, $\prescript{I}{}{\boldsymbol{T}}^{k-1}_{k}=[R^{k-1}_{k}|p^{k-1}_{k}]$, can be computed. We transform the relative transformation into camera coordinate using an extrinsic parameter, $\prescript{C}{I}{\boldsymbol{T}}$, between camera and the IMU sensor. We can obtain the good intial guess, $\prescript{C}{}{\boldsymbol{T}}_{k}$, for GICP tracking from the IMU preintegration.
\begin{align}
\prescript{C}{}{\boldsymbol{T}}^{k-1}_{k} &= \prescript{C}{I}{\boldsymbol{T}} \prescript{I}{}{\boldsymbol{T}}^{k-1}_{k} \\
\prescript{C}{}{\boldsymbol{T}}_{k} &= \prescript{C}{}{\boldsymbol{T}}^{k-1}_{k} \prescript{C}{}{\boldsymbol{T}}_{k-1}.
\end{align}

\subsection{Update Preintegration}
IMU-based GICP tracking allows for faster and more reliable performance compared to the original GICP method. This is because it leverages the accurate initial guess provided by the IMU, rather than estimating the initial guess based on the previous frame's position or using a constant velocity model. 
IMU preintegration continuously integrates accelerations and angular velocities over time. However, due to the fact that the IMU sensor captures continuous measurements, drift accumulates at low speeds, which results in a degradation of the tracking performance over time caused by the accumulation of integration errors.
To address such degradation, it is necessary to update the initial values of the IMU preintegration parameters—position $p^{k}_{t}$, velocity $v^{k}_{t}$, and rotation $R^{k}_{t}$—using the optimized transformation $\prescript{C}{}{\boldsymbol{T}}^*_{k}=[\prescript{C}{}{R}^*_k|\prescript{C}{}{p}^*_k]$ obtained from GICP tracking.
\begin{equation}
\prescript{I}{}{\boldsymbol{T}}^*_{k}=\prescript{I}{C}{\boldsymbol{T}}\prescript{C}{}{\boldsymbol{T}}^*_{k},
\end{equation}
\begin{equation}
    p^{k}_{t} \gets \prescript{I}{}{p}^*_{k}, R^{k}_{t} \gets \prescript{I}{}{R}^*_{k}, v^{k}_{t} \gets \frac{\prescript{I}{}{p}^*_k - \prescript{I}{}{p}^*_{k-1}}{\Delta t},
\end{equation}
where $\Delta t$ represents the time interval between consecutive RGB-D frames, and $\prescript{C}{}{p}^{*}_{k}$ and $\prescript{C}{}{R}^{*}_{k}$ denote the results obtained from GICP tracking for position and rotation, respectively.

\begin{sloppypar}
\begin{table*}[ht]
\caption{Tracking and Rendering Performance compared to Existing Methods on uHumansV1 Dataset}
\centering
\scriptsize % reduce font size
\setlength{\tabcolsep}{2pt} % reduce column separation
\begin{tabular}{@{}cccccccccccccc@{}}
\toprule
\multicolumn{2}{c}{\multirow{2}{*}{uHumansV1\cite{Rosinol20rss-dynamicSceneGraphs_uhumans1}}} & 
\multicolumn{4}{c}{\textit{\begin{tabular}[c]{@{}c@{}}Humans12\end{tabular}}} & 
\multicolumn{4}{c}{\textit{\begin{tabular}[c]{@{}c@{}}Humans24\end{tabular}}} & 
\multicolumn{4}{c}{\textit{\begin{tabular}[c]{@{}c@{}}Humans60\end{tabular}}} \\ \cmidrule(l){3-14} 
& & ATE $\downarrow$ (cm) & PSNR $\uparrow$ & SSIM $\uparrow$ & LPIPS$\downarrow$ & ATE $\downarrow$ (cm) & PSNR $\uparrow$ & SSIM $\uparrow$ & LPIPS $\downarrow$ & ATE $\downarrow$ (cm) & PSNR $\uparrow$ & SSIM $\uparrow$ & LPIPS $\downarrow$ \\ \midrule
\multirow{1}{*}{Traditional}& ORB-SLAM3\cite{mur2017orb} & 191.76 & - & - & - & 15.82 & - & - & - & 17.15 & - & - & -  \\
\multirow{1}{*}{VIO}& VINS-Mono\cite{qin2018vins} & 18.85 & - & - & - & 25.60 & - & - & - & 22.29 & - & - & - \\ \midrule
\multirow{3}{*}{3DGS} & MonoGS\cite{matsuki2024monogs}  & 603.50 & 20.99 & 0.714 & 0.503 & 1138.15 & 21.35 & 0.740 & 0.499 & 970.68 & 17.49 & 0.637 & 0.612 \\
& PhotoSLAM\cite{huang2024photoslam}  & 650.97 & 14.27 & 0.596 & 0.598 & 1573.21 & 14.23 & 0.609 & 0.592 & 1315.79 & 14.78 & 0.596 & 0.587 \\
& GS-ICP SLAM\cite{ha2024gsicpslam} & 677.87 & 22.89 & 0.788 & 0.345 & 776.79 & 23.49 & 0.794 & 0.334 & 973.49 & 20.85 & 0.741 & 0.422 \\\midrule
& Ours (VIGS SLAM) & \textbf{35.35} & \textbf{26.22} & \textbf{0.849} & \textbf{0.205} & \textbf{25.03} & \textbf{25.89} & \textbf{0.853} & \textbf{0.207} & \textbf{46.86} & \textbf{23.91} & \textbf{0.820} & \textbf{0.252} \\ \bottomrule
\end{tabular}
\label{table1}
\begin{tablenotes}
\item {\hspace{0.6cm}The algorithm with the highest performance is indicated in \textbf{bold}, and the second-best is \underline{underlined}. Traditional VIO methods are excepted in best algorithm. }
\end{tablenotes}
\end{table*}
\end{sloppypar}

\begin{figure*}[h]
\centering
\includegraphics[width=\textwidth]{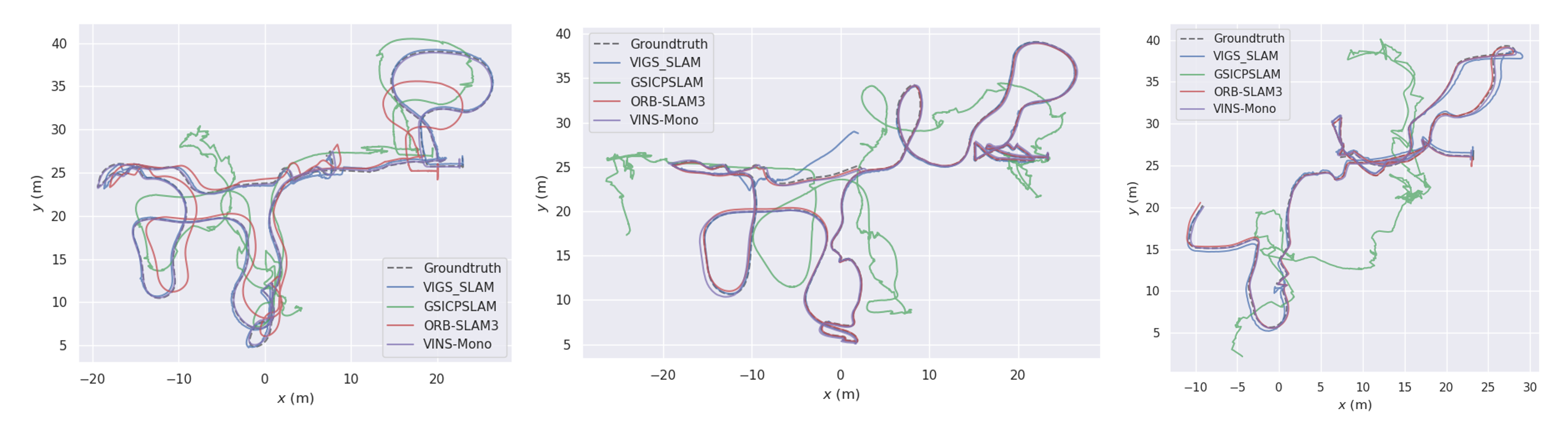}
\caption{ Trajectory in Humans12(left), Humans24(middle), Humans60(right) of uHumansV1, compared with ORB-SLAM3, VINS-Mono, GS-ICP SLAM, and VIGS SLAM(ours).}
\label{fig_trajectory_uhumansv1}
\end{figure*}

\subsection{3D Gaussian Splatting Mapping}
In this paper, the map representation is based on 3D Gaussian Splatting\cite{kerbl20233dgs}, using a set of Gaussian models \textit{G} to represent the 3D space.
Each Gaussian $\mathit{G}_{m}$ is defined by its RGB color $c_{m}$, opacity $\sigma_{m}$, center position $\mu_{m}$, and scale $s_i$. 
As mentioned in the section \ref{GICP tracking}, the input point cloud for the GICP tracking part is expressed as a Gaussian distribution $\mathit{G}_{m}$, with its center $\mu_{m}$ and covariance $\Sigma_{m}$ calculated accordingly. 
These values are then reused in the mapping process, eliminating the need for redundant computations.
\begin{align}
& {G}_{m}(x) = \text{Exp}(-{{1}\over{2}}(x - \mu_{m})\Sigma_{m}^{-1}(x - \mu_{m})^{T})
\end{align}

The covariance can be expressed through singular value decomposition as $\Sigma_{m}={R}{\Lambda}^{2}{R}^{T}$, where $R$ represents the orientation of the Gaussian, and $\Lambda=diag(s_2,s_1,s_0)$ denotes the scale matrix. 
The map represented using the 3DGS method is rendered into a 2D image through volume rendering, where the color values of the pixels in the image are determined by the contributions of the $\mathcal{N}$ Gaussians that make up the pixel $p$.

The color of a pixel $C_p$ can be computed using the following equation:
\begin{align}
& {C}_{p} = \sum_{{m\in \mathcal{N}}}{c_{m}\alpha_{m}\prod_{n=1}^{m-1}{(1-\alpha_{n}})},
\end{align}
where $c_i$ denotes the color of the $i$th Gaussian, and $\alpha_i$ represents the opacity sampled from the $i$th Gaussian distribution at the pixel position. 
Similarly, the opacity at a pixel position can be calculated as follows:
\begin{align}
& {O}_{p} = \sum_{{m\in \mathcal{N}}}{\alpha_{m}\prod_{n=1}^{m-1}{(1-\alpha_{n}})}
\end{align}

To optimize the Gaussians representing the rendered images, we utilize the dense photometric loss ${L}_{photo}$, which is the L1 loss between the rendered and observed images, the SSIM image loss $\mathcal{L}_{SSIM}$, and the depth loss $\mathcal{L}_{depth}$, which is the L1 loss between the rendered and observed depth images. 
The combined mapping loss function is given by:
\begin{align}
& \mathcal{L}_{mapping} = (1-\lambda_I)\mathcal{L}_{photo} + \lambda_I\mathcal{L}_{SSIM} + \lambda_D\mathcal{L}_{depth}
\end{align}

While we employ the same loss functions as the original 3DGS for mapping, unlike traditional 3DGS, where all Gaussians are optimized simultaneously, SLAM tasks progressively introduce new Gaussians. To prevent scale imbalance between Gaussians introduced early and those added later in the process, we apply scale normalization \cite{ha2024gsicpslam}.

\section{EXPERIMENTS}
\label{experiments}
The proposed method was evaluated using the photo-realistic and large-scale visual-inertial datasets uHumansV1\cite{Rosinol20rss-dynamicSceneGraphs_uhumans1} and uHumansV2\cite{Rosinol21ijrr-Kimera_uhumans2}. 
The uHumansV1 dataset was collected in a $65m\times65m$ office space, with each scenario containing varying numbers of people, specifically 12, 24, and 60 individuals. 
This dataset is well-suited for visual inertial odometry and visual inertial SLAM, although it has the limitation of providing only grayscale image data.
We also conducted experiments on the uHumansV2 dataset ranging from small apartment scenes to large office scenes, to evaluate the mapping and rendering performance of the proposed algorithm given RGB images.

\subsection{Experiments Setup}
All experiments were conducted on a desktop equipped with an Intel i5-12500 CPU, 32GB of RAM, and an NVIDIA RTX A5000 GPU with 24GB of memory. 
The accuracy of tracking was evaluated using the Root Mean Square Error (RMSE) of the Absolute Trajectory Error (ATE)\cite{grupp2017evo}. 
To assess the mapping and rendering processes, metrics such as Peak Signal-to-Noise Ratio (PSNR), Structural Similarity Index Measure (SSIM), and Learned Perceptual Image Patch Similarity (LPIPS) were utilized. 
The methods used for comparison include the recently proposed 3DGS SLAM approaches such as MonoGS\cite{matsuki2024monogs} and PhotoSLAM\cite{huang2024photoslam} as well as ORB-SLAM3\cite{campos2021orb} and VINS-Mono\cite{qin2018vins}. 
In addition, GS-ICP SLAM\cite{ha2024gsicpslam} was included as a baseline.

\subsection{Performance on the uHumansV1}

\begin{figure*}[t]
\centering
\includegraphics[width=0.98\textwidth]{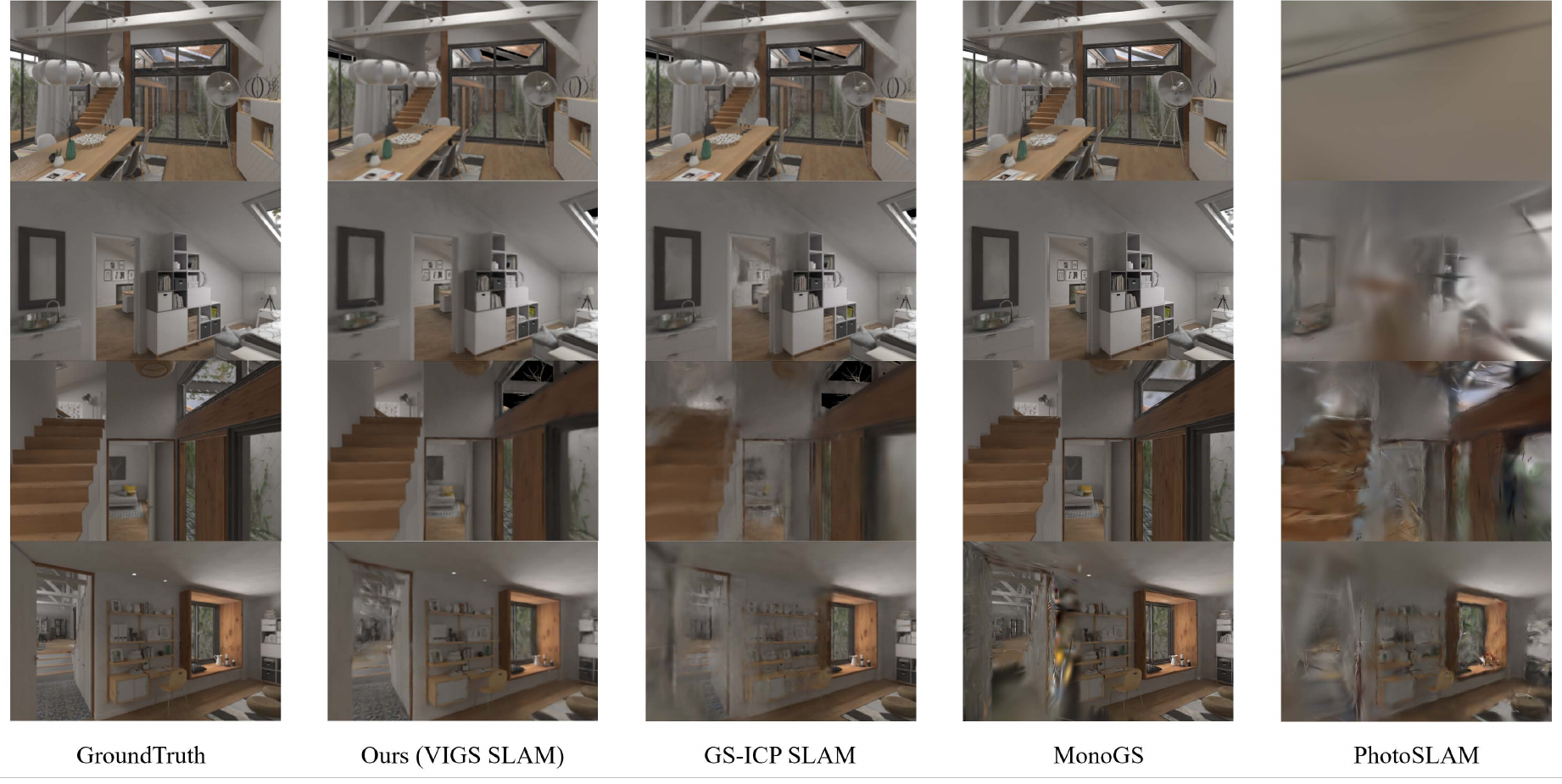}
\caption{Qualitative rendering results on the apartment scene of uHumansV2 dataset, compared with VIGS SLAM(ours), GS-ICP SLAM, MonoGS, and PhotoSLAM.}
\label{fig_rendering_image_uhumansv2}
\end{figure*}

To evaluate the performance of our method compared to state-of-the-art method, we conducted experiments on uHumansV1, as detailed results are shown in Tab. \ref{table1}.

\noindent \textbf{Tracking Performance.} 
VIGS SLAM demonstrates superior tracking performance compared to existing 3DGS SLAM approaches, achieving tracking performance on par with traditional VIO methods, as shown in Fig.\ref{fig_trajectory_uhumansv1}. These differences stem from the lower similarity between consecutive images in large-scale environments with dynamic objects. While existing 3DGS SLAM methods are effective in room-scale static datasets, the uHumansV1 dataset presents a greater challenge due to its larger spatial coverage over the same time frame, reducing the similarity between consecutive images. Additionally, 3DGS methods that rely on dense photometric loss for tracking are more vulnerable to dynamic objects, thus limiting their performance. Compared with GS-ICP SLAM, VIGS SLAM outperforms in all scenarios and significantly reduces the tracking error from a few meters to several tens of centimeters. These results indicate that by employing IMU preintegration values as the initial guess, our method achieves a notable tracking accuracy.

\noindent \textbf{Mapping \& Rendering Performance.} 
As demonstrated in Tab. \ref{table1}, the VIGS SLAM achieves the highest rendering performance. 
Since MonoGS optimize the dense photometric loss for tracking and PhotoSLAM utilize the sparse feature based tracking, poor tracking performance of MonoGS and PhotoSLAM leads to poor rendering performance.
GS-ICP SLAM shows relatively high rendering performance, but the mapping performance deteriorates due to poor tracking performance in the revisited area, as shown in Fig. \ref{fig_mapping_uhumansv1} and Fig. \ref{fig_trajectory_uhumansv1}.
Fig. \ref{fig_rendering_image_uhumansv1} and \ref{fig_mapping_uhumansv1} highlight the superior rendering and mapping performance of the proposed method compared to GS-ICP SLAM. In particular, VIGS SLAM successfully renders images with humans cleanly excluded. This result indicates that the integration of IMU sensor enables robust correspondences even in the presence of dynamic objects.

\subsection{Performance on the uHumansV2}

As reported in Tab. \ref{table2} and Tab. \ref{table3}, the results on the uHumansV2 dataset demonstrate that our method also performs well in environments where RGB images are provided.
While MonoGS and GS-ICP SLAM demonstrate high tracking accuracy in the small-scale apartment scene, they notably degrade in the large-scale office scene. 
Although PhotoSLAM achieves low translation errors in the office environment, it suffers from decoupling between front-end and back-end, leading to poor overall mapping performance in both scenes. 
In contrast, our method shows consistently robust tracking and rendering performance across both environments. 
\begin{sloppypar}
\begin{table}[t]
\centering
\caption{Tracking Performance compared to Existing Methods on uHumansV2 Dataset (ATE RMSE is in cm)}
\scriptsize % reduce font size
\setlength{\tabcolsep}{2pt} % reduce column separation
\begin{tabular}{@{}cccc@{}}
\toprule
\multicolumn{2}{c}{uHumansV2\cite{Rosinol21ijrr-Kimera_uhumans2}} & 
{\textit{\begin{tabular}[c]{@{}c@{}}Apartment\end{tabular}}} & 
{\textit{\begin{tabular}[c]{@{}c@{}}Office\end{tabular}}}\\ \midrule
\multirow{3}{*}{3DGS} & MonoGS\cite{matsuki2024monogs}  & \underline{16.64} & 802.86 \\
& PhotoSLAM\cite{huang2024photoslam}  & 285.99 & \textbf{34.74} \\
& GS-ICP SLAM \cite{ha2024gsicpslam}& 29.21 & 674.38 \\ \midrule
& Ours (VIGS SLAM) & \textbf{16.17} & \underline{144.72} \\ \bottomrule
\end{tabular}
\label{table2}
\begin{tablenotes}
\item {The best-performing algorithms are shown in \textbf{bold}, and second best is \underline{underlined}}
\end{tablenotes}
\end{table}
\end{sloppypar}

Fig. \ref{fig_rendering_image_uhumansv2} illustrates a qualitative rendering results from our method and those of the comparison methods. 
Most methods fail to accurately render the decoration on the right shelf because they are too thin in the image in the first row, but our method successfully captures these details.
In the third row, methods with poor tracking performance produce maps with incorrect poses, resulting in low-quality rendered image. In contrast, our method generates high-quality image from accurate poses. The last row corresponds to the end part of the sequence, where less  observations were made around the door. Despite strong tracking performance of MonoGS, it renders a low-quality image in this area. On the other hand, VIGS SLAM, which utilizes the dense point cloud matching, maintains robust rendering quality.

\begin{sloppypar}
\begin{table}[t]
\caption{Rendering Performance compared to Existing Methods on uHumansV2 Dataset}
\scriptsize % reduce font size
\setlength{\tabcolsep}{2pt} % reduce column separation
\begin{tabular}{@{}cccccccc@{}}
\toprule
\multicolumn{2}{c}{\multirow{2}{*}{uHumansV2\cite{Rosinol21ijrr-Kimera_uhumans2}}} & 
\multicolumn{3}{c}{\textit{\begin{tabular}[c]{@{}c@{}}Apartment\end{tabular}}} & 
\multicolumn{3}{c}{\textit{\begin{tabular}[c]{@{}c@{}}Office\end{tabular}}}\\ \cmidrule(l){3-8} 
&  & PSNR $\uparrow$ & SSIM $\uparrow$ & LPIPS$\downarrow$ & PSNR $\uparrow$ & SSIM $\uparrow$ & LPIPS $\downarrow$  \\ \midrule
\multirow{3}{*}{3DGS} & MonoGS\cite{matsuki2024monogs} & \textbf{29.68} & \textbf{0.883} & \textbf{0.196} & 20.56 & 0.696 & 0.596  \\
& PhotoSLAM\cite{huang2024photoslam} & 19.13 & 0.691 & 0.481 & 14.35 & 0.585 & 0.646 \\
& GS-ICP SLAM \cite{ha2024gsicpslam} & 26.54 & 0.813 & 0.333 & \underline{22.92} & \underline{0.773} & \underline{0.378} \\ \midrule
& Ours (VIGS SLAM) & \underline{27.26} & \underline{0.826} & \underline{0.289} & \textbf{23.88} & \textbf{0.793} & \textbf{0.330}\\ \bottomrule
\end{tabular}
\label{table3}
\begin{tablenotes}
\item {The best-performing algorithms are shown in \textbf{bold}, and second best is \underline{underlined}}
\end{tablenotes}
\end{table}
\end{sloppypar}

\section{CONCLUSIONS \& FUTURE WORK}
\label{conclusions}
In this paper, we proposed VIGS SLAM, a novel framework that leverages visual, depth, and inertial measurements to achieve improved tracking performance and 3D map reconstruction in large-scale environments. To evaluate the effectiveness of our approach, we conducted experiments on the uHumansV1 and uHumansV2 datasets, which provide large-scale visual-inertial data. As a result, we demonstrate that our method outperforms SOTA methods through tracking and mapping performance comparisons in most datasets.

As future work, we intend  to develop a tightly-coupled visual-inertial framework. Furthermore, we will integrate a loop-closing module to address long-term drift, thereby ensuring more robust and accurate SLAM performance.

\nocite{*}

\bibliographystyle{IEEEtran}
\bibliography{bib}

\end{document}